\documentclass[conference]{IEEEtran}
\IEEEoverridecommandlockouts
\usepackage{cite}
\usepackage{amsmath,amssymb,amsfonts}
\usepackage{algorithmic}
\usepackage{graphicx}
\usepackage{textcomp}
\usepackage{xcolor}
\usepackage{booktabs}
\usepackage{bbm}
\usepackage{multirow} 

\usepackage[caption=false,font=footnotesize]{subfig}

\graphicspath{{./images/}}

\usepackage{geometry}
\usepackage[hidelinks]{hyperref}

\def\BibTeX{{\rm B\kern-.05em{\sc i\kern-.025em b}\kern-.08em
    T\kern-.1667em\lower.7ex\hbox{E}\kern-.125emX}}

\begin{document}

\title{Trans-XFed: An Explainable Federated Learning for Supply Chain Credit Assessment\\
}


\author{
    \IEEEauthorblockN{Jie Shi, Arno P. J. M. Siebes, Siamak Mehrkanoon}
    \IEEEauthorblockA{
        Department of Information and Computing Sciences, Utrecht University\\
        Utrecht, The Netherlands\\
        j.shi1@uu.nl, a.p.j.m.siebes@uu.nl, s.mehrkanoon@uu.nl
    }
}

\maketitle

\begin{abstract}
This paper proposes a Trans-XFed architecture that combines federated learning with explainable AI techniques for supply chain credit assessment. The proposed model aims to address several key challenges, including privacy, information silos, class imbalance, non-identically and independently distributed (Non-IID) data, and model interpretability in supply chain credit assessment. We introduce a performance-based client selection strategy (PBCS) to tackle class imbalance and Non-IID problems. This strategy achieves faster convergence by selecting clients with higher local F1 scores. The FedProx architecture, enhanced with homomorphic encryption, is used as the core model, and further incorporates a transformer encoder. The transformer encoder block provides insights into the learned features. Additionally, we employ the integrated gradient explainable AI technique to offer insights into decision-making. We demonstrate the effectiveness of Trans-XFed through experimental evaluations on real-world supply chain datasets. The obtained results show its ability to deliver accurate credit assessments compared to several baselines, while maintaining transparency and privacy. The code is available on GitHub\footnote{https://github.com/JieJieNiu/Trans-XFed}.
\end{abstract}


\footnotetext{Accepted by IEEE FLTA 2025. This is the author’s preprint.}

\section{Introduction}
Financial institutions often have access to limited supply chain datasets, which may not adequately represent the market's diverse range of supply chain credit profiles \cite{moretto2019supply}. Due to the customer marketing strategies employed by these institutions, historical data on supply chain loans tends to be concentrated within specific industries or customer segments. Furthermore, the entirety of the market's supply chain credit data is likely dispersed across multiple financial institutions. However, with increasing competition among financial institutions and rising concerns over user privacy and data security, it has become challenging for data providers to share their data.


Federated learning (FL) enables distributed ML across decentralized data without compromising client privacy \cite{hu2021federated}. Federated learning enables the development of models trained on more representative datasets, helping to reduce bias and promote fairness in credit assessment decisions. It allows financial institutions to collaborate with other data providers to train credit assessment models without centralizing sensitive customer data. This access to diverse data sources improves the robustness and accuracy of credit assessment models while achieving the goal of \lq available and invisible\rq data sharing. Although the model parameters and gradients do not directly include raw client data, they can expose patterns or distributions related to the underlying datasets \cite{zhu2019deep}. Inference attacks \cite{nasr2019comprehensive} might exploit these patterns to deduce sensitive information about the clients. Homomorphic encryption \cite{yousuf2020systematic} ensures that even these sensitive insights remain secure during parameter exchange.

While federated learning (FL) offers a promising approach to enhancing privacy protection, it faces challenges when applied in real-world scenarios compared to centralized learning. In the context of supply chain credit, FL encounters specific difficulties, including Non-IID (Non-Independent and Identically Distributed) data distributions \cite{10361408,10840118,10839814}, class imbalance \cite{liu2020alike}, and interpretability issues associated with black-box models \cite{hassija2024interpreting}. The nature of supply chain credit data often results in Non-IID distributions, where data samples may vary significantly across industries or time periods, complicating the use of traditional ML algorithms. Class imbalance presents another challenge, as the skewed distribution of defaulting and non-defaulting instances can lead to biased model predictions and reduced performance. Additionally, the reliance on sophisticated, often opaque ML algorithms (black-box models) makes it difficult to interpret model predictions, potentially hindering regulatory compliance and eroding stakeholder trust. Finally, the sensitive nature of credit data requires that both data and models are securely managed throughout the sharing process.

In this paper, we propose a novel performance-based client selection strategy, adapted within a Trans-XFed model built upon a core federated learning framework. Additionally, we use explainable AI techniques to enhance the transparency and interpretability of supply chain credit assessment. Our work offers four distinct contributions:
\begin{itemize} 
\item To address issues of data isolation and privacy, we adopt Federated Learning with a Homomorphic Encryption privacy technique \cite{he}. This centralized credit assessment model can handle the diversity and distribution of supply chain data, enabling different financial institutions to collaboratively train models across decentralized data sources without sharing sensitive information.

\item To solve the Non-IID problem, we introduce a performance-based client selection strategy (PBCS) built on the FedProx architecture \cite{fedprox}. PBCS enhances communication efficiency by allowing the central server to dynamically select clients based on their local model performance. FedProx introduces a proximal term that encourages local models to remain close to their local updates while contributing to the global model, reducing communication costs and enabling faster convergence for improved efficiency.

\item To resolve the class imbalance issue, we utilize a weighted Negative Log-likelihood loss function \cite{shimodaira2000improving}, assigning a higher weight to defaulting samples. This approach significantly improves recall and F1 scores for the minority class.

\item To address the black-box problem, we provide feature-to-feature interpretability through an attention mechanism \cite{vaswani2017attention} and local feature-to-output explainability using integrated gradients XAI techniques \cite{ig}. The self-attention mechanism extracts relationships among features, while integrated gradients retain the gradients locally, thereby preserving privacy by preventing any data sharing that could compromise it. \end{itemize}


\begin{figure*}[!htbp]
\centering
{\includegraphics[width=17cm]{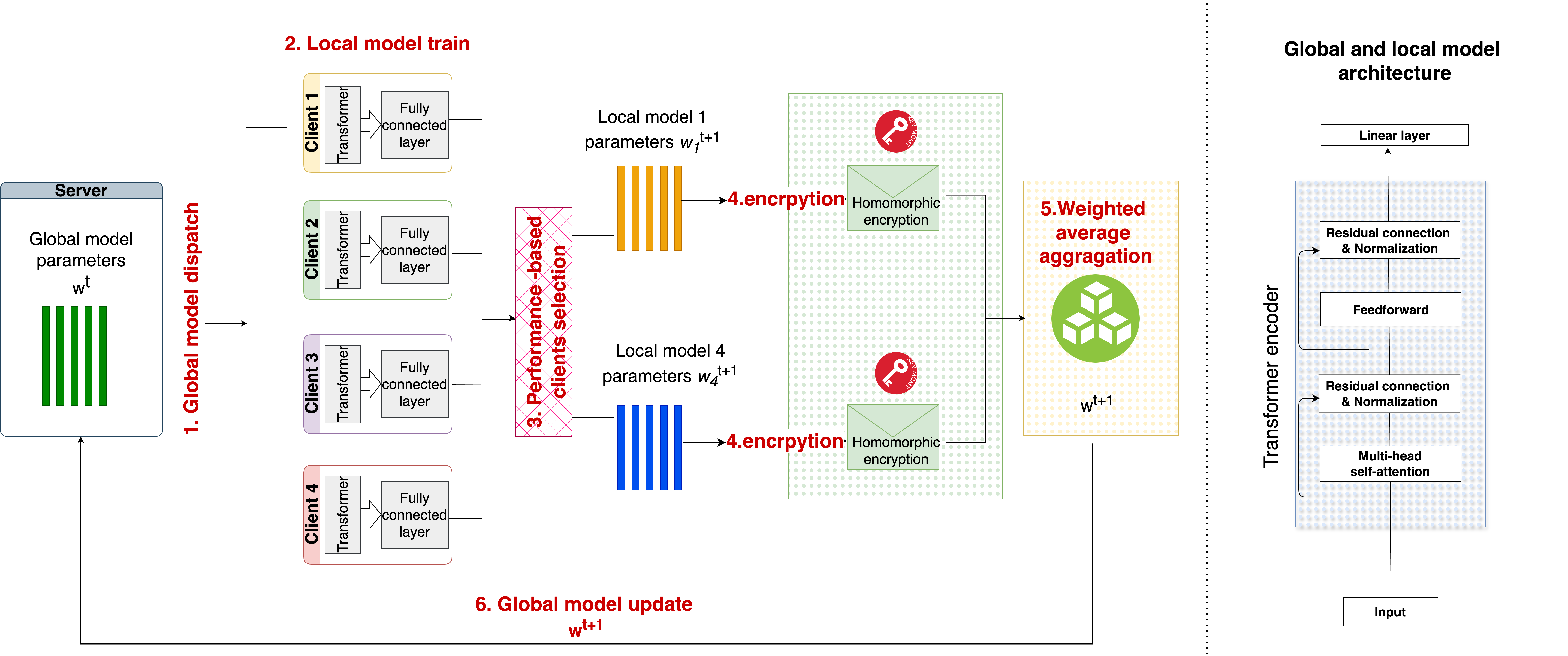}}
\caption{The architecture of Trans-XFed. This graph depicts the overall framework of our proposed model. This is an example of clients 1 and 4 are chosen for participants in t+1 communication rounds.}\label{fig1}
\end{figure*}

\section{Related Work}\label{section:sec2}
\subsection{Federated learning in credit assessment}
Federated learning is an emerging machine learning technique that is increasingly utilized in the financial sector, particularly in the context of credit assessment.Zheng et al. \cite{zheng2020vertical} developed a vertical FL method for interpretable scorecards, emphasizing data privacy. Li et al. \cite{li2023research} reported a 14\% improvement in managing credit risk for small banks using FL. Zhang et al. \cite{zhang2024effects} showed FL improves performance for smaller datasets, with a 17.92\% average improvement, and suggested incentives for dominant client participation. Yang et al. \cite{yang2022privacy} proposed combining FL with blockchain for secure and explainable credit scoring.

\subsection{Client Selection in FL}
Client selection in federated learning is a crucial aspect that impacts the performance and efficiency of the training process. FedCS \cite{nishio2019client} is a client selection framework that addresses resource constraints to aggregate client updates and accelerate performance improvement in machine learning models. Cho et al. \cite{powerofchoice} conduct a convergence analysis of biased client selection strategies, highlighting the importance of biasing towards clients with highest local F1 scores for faster error convergence. GREED \cite{albelaihi2022green} is an energy-aware client selection method that can optimize the trade-off between maximizing selected clients and minimizing energy consumption. Furthermore, Yang et al. \cite{9632344} present a client selection framework, which enhances the efficiency of training high-performance, privacy-preserving machine learning models by optimizing client selection based on resource conditions, thereby accelerating the training process in heterogeneous networks. A recent study proposes Bayesian Federated Learning with Stochastic Variational Inference (BayFL-SVI) \cite{10840014} to address non-IID data and aggregation inefficiency, leveraging client-wise ELBO computation to weight updates, thereby improving convergence speed, model accuracy, and robustness with theoretical guarantees.

\section{Proposed Approach}

\subsection{Problem definition}

In the cross-device federated learning, we examine total of $K$ clients, each client $k\in[K]$ has a local dataset $D_k$. The input of each dataset $X_{k}=\left\{x^{(i)}_{k}\right\}_{i=1}^{N_k}$, where  $x_{k}^{(i)} \in \mathbb{R}^{d}$, output $Y_{k}=\left\{y^{(i)}_{k}\right\}_{i=1}^{N_k}$. Here, $N_k$ represents the sample size for each clients, and the datasets of each clients are of the same dimension $d$. This research belongs to horizontal federated learning, as each clients has a different set of samples from the same feature space. These clients are linked through a central aggregation server, aiming to collaboratively find the model parameter that reduces the empirical risk. The client's sample size $N_k$ varies significantly across clients. Meanwhile, there is an imbalance between positive and negative samples across different clients. 

\subsection{Overall architecture}

Our proposed model builds upon and extends the FedProx architecture \cite{li2020federated}. We proposed a performance-based clients selection as our clients selection strategy (for more details see section \ref{seca}). To ensure parameter security, we utilize the CKKS (Cheon-Kim-Kim-Song) homomorphic encryption scheme \cite{cheon2017homomorphic}. The Negative Log-Likelihood (NLL) loss function is employed to address class imbalance, and the Integrated Gradients method \cite{sundararajan2016gradients} is applied for result interpretability. In the following, we will explain the training workflow and each element of the proposed model.

\subsubsection{Federated Training Workflow in Trans-XFed}
 The diagram summarizing our proposed explainable transformer FedProx networks (Trans-XFed), is depicted in Fig.\ref{fig1}. The algorithm divides the training into communication rounds $t, t+1,...t+\tau$, where each round consists of the following six steps:

\begin{enumerate}
\item \textbf{Global model dispatch}: In this first step, the central server initializes a global model with random weights $w^{t}$, and assigns it to the local models $w_k^{t}$, for $k = 1,...,K$. This model parameter serves as the starting point for training across all client devices.
\item \textbf{Client selection}: We adopt a client selection strategy called performance-based client selection to select a subset of M clients with highest local F1 score to participant in the next round of training, aiming to achieve faster convergence.
\item \textbf{Local model training}: All selected M clients train the model locally on their own datasets without sharing the raw data. The local training can involve multiple epochs to improve the model's performance on the local data.
\item \textbf{Parameters encryption}: In this step, all the selected model's parameters are encrypted using fully homomorphic encryption to ensure privacy.
\item \textbf{Parameters aggregation}: The central server collects the locally encrypted model parameters from all participating clients and aggregates these parameters to compute a global update $w^{t+1}$. This aggregation is performed using a weighted average based on the sample size of each selected client.
\item \textbf{Global model update}: Apply the aggregated global update to the global model. This step updates the global model's weights, integrating the parameters from all participating clients during the current round.
\end{enumerate} 
The model repeats steps 1-6 for multiple communication rounds until convergence.

\subsubsection{FedProx}
The critical innovation of FedProx \cite{li2020federated} lies in incorporating a proximal term $\frac{\mu}{2}\left\|w_k^{t+1}-w^{t}\right\|^{2}$ into the objective function during model aggregation to ensure that the updated global model remains close to the previous global model. This proximal term penalizes the discrepancy between the local and global model parameters, encouraging local updates to align more closely with the global model. It helps to prevent drastic changes and ensures more stable convergence. The central server coordinates the global learning objective across the network. It aims to minimize local function $F_{k}(w_k^{t+1})$ for client $k$, which typically quantifies the loss of the model on the data held by client $k$ based on the updated parameters $w_k^{t+1}$. Meanwhile, client $k$ uses its local solver of choice to approximately minimize the following objective function $g_{k}$ \cite{fedprox}: 

\begin{equation}
\min _{w_k} g_{k}\left(w_k^{t+1} ; w^{t}\right)=F_{k}(w_k^{t+1})+\frac{\mu}{2}\left\|w_k^{t+1}-w^{t}\right\|^{2},
\end{equation}
where $\mu$ is the proximal term coefficient that controls how strongly the local model is penalized for deviating from the global model, and $\left\| \right\|^{2}$ is the Euclidean norm of local parameters $w_k^{t+1}$ and global parameters $w^{t}$.

\subsubsection{Local and global model}

The local and global models share the same architecture, consisting of a transformer encoder block \cite{vaswani2017attention} and fully connected layers. The transformer encoder processes the input $X_k$, generating new embeddings through self-attention and feed-forward layers, complemented by layer normalization and residual connections. This design effectively captures information and creates rich representations of the input. The primary component of the transformer encoder is the multi-head self-attention mechanism \cite{vaswani2017attention}, where each attention head computes scores for each feature relative to all other features.

\subsubsection{Loss function}


We employ the weighted Negative Log-likelihood (NLL) loss function as $F_{k}(w_k^{t})$ to optimize our model. The NLL loss \cite{lastras2019information} is commonly used in classification tasks, mainly when dealing with locally imbalanced datasets \cite{shimodaira2000improving}. In weighted NLL loss, each class is assigned a weight that reflects its importance. These weights are typically inversely proportional to the frequency of each class in the dataset. In this research, the defaulting class is rare and is assigned a higher weight to give it more emphasis during training. Given a weight $\beta_{c}$, the weighted NLL loss for client $k$ can be written as follows \cite{shimodaira2000improving}: 

\begin{equation}
\begin{aligned}
\ell_{NLL}=-\sum_{i=1}^{N_k} \sum_{c=1}^{C}\beta_c y_{k,c}^{(i)}\log P\left(y_{k,c}^{(i)} \mid x_{k,c}^{(i)}\right),
\end{aligned}
\end{equation}
where $N$ is the number of samples, $x_{k,c}^{(i)}$ represents the input instance $i$ of class $c$, $y_{k,c}^{(i)}$ denotes the true class for instance $i$ in class $c$, $\beta_c$ is the weight assign to the class $c$, $\log P$ is the logarithm of the predicted probability for the true class.

\subsubsection{Clients selection}\label{seca}

At each communication round, instead of randomly sampling clients, we partially select clients with highest F1 scores to participate in the training (performance-based client selection strategy). The server sends the current global model parameters $w^t$ to $K$ clients and ranks them based on their local F1 scores. The top $M$ clients with the highest F1 scores are selected, where $M = rK$ and $r$ represents the selection ratio. The set $S$ consists of the indices of the selected $M$ clients. Next, only the $M$ selected clients will get trained during the next communication round. Then, the $M$ clients send their locally trained model parameters to the server. The server aggregates these parameters and updates the global model parameters to $w^{t+1}$.

By selecting clients with highest F1 scores, the model can learn from clients who provide more accurate and balanced data, potentially improving the global model’s overall performance faster than random selection. Moreover, leveraging clients with highest F1 scores may lead to faster convergence as these clients likely contribute higher-quality updates, making the model more stable and reducing the number of rounds needed for effective training.

\subsubsection{Weighted aggragation}
Fully homomorphic encryption (FHE) \cite{yousuf2020systematic} allows computations to be carried out directly on encrypted parameters $E_{pk}(w_k^{t+1})$ without needing to decrypt it, preserving its privacy throughout the aggregation process. 
The CKKS (Cheon-Kim-Kim-Song) \cite{cheon2017homomorphic} is an FHE scheme. Unlike traditional FHE that support exact arithmetic, it supports approximate arithmetic for real numbers. The decrypted results are approximate numbers compared to the plaintext, which is acceptable for machine learning. It is considered anti-quantum secure, meaning it can withstand attacks from classical and quantum computers. 

The server initializes the CKKS parameters and generates public key and secret key. After that, the public and secret keys are sent to all participants. With this, the model can encrypt the plaintext data $w_k$ using a public key, resulting in ciphertexts $E_{pk}(w_k^{t+1})$.  For a given aggregation weighing factor $\gamma_{k}$, the server aggregates the chosen models' parameters as follows \cite{cheon2017homomorphic}:

\begin{equation}
\begin{split}
w^{t+1} = D_{sk}\left( \sum_{k\in \left \{ S \right \} } \gamma _{k}E_{pk}(w_{k}^{t+1}) \right), 
\end{split}
\end{equation}
where $D_{sk}$ and $E_{pk}$ are the decryption and encryption function describe in \cite{cheon2017homomorphic}. $\left \{ S \right \}$ is the set of the indices of the selected $M$ clients, defined as previously.

\subsection{Interpreting the predictions}
Integrated Gradients is a method used in machine learning to explain the predictions of deep learning models \cite{sundararajan2016gradients}. In Fed-XAI, its application can reduce the risks of exposing users’ privacy to a certain extent, as both the data used and the gradients remain on the client side. It is an attribution method used to interpret deep neural networks by assigning importance scores to each input feature, illustrating their contribution to a model's prediction. This method calculates the gradients of the model's output with respect to input features along a path from a baseline. The baseline typically is an input with all features set to zero. The integrated gradient for a feature is computed by summing the gradients of the model's output with respect to the input as it varies from the baseline to the input, averaged over a predefined number of steps.

\begin{table*}[ht]
\centering
\normalsize
\caption{The best recall, precision, and F1 scores of the minority class (defaulting) on testing datasets from 50 communication rounds using the Trans-XFed and other examined models. The communication round column represents the round in which the best model appears.}\label{tbl1}

\begin{tabular}{cccccc}
\toprule[1pt]
\textbf{Fed models} & \textbf{loss} & \textbf{{best recall}$\uparrow$} & \textbf{{best precision}$\uparrow$} & \textbf{{best F1 scores}$\uparrow$} & \textbf{{communication round} $\downarrow$}\\ 
\toprule[1pt]
 & CE & 0.8853 & 0.5883 & 0.7069 & 20 \\
\text{Trans-XFed} & Focal & 0.6059 & \underline{0.7102} & 0.6534 & 13\\
& NLL & \underline{0.8889} & 0.5961 & \underline{0.7137} & \underline{10} \\
&&&&&\\
 & CE & 0.8884 & 0.5234 & 0.6588 & 42 \\
\text{FedProX}\cite{fedprox} & Focal & 0.6918 &0.4180 & 0.5207 & 34 \\
& NLL & 0.8486 & 0.5456 & 0.6642 & 36\\
&&&&&\\
 & CE & 0.8719 & 0.5916& 0.7049 & 38 \\
\text{FedAvg}\cite{mcmahan2017communication} & Focal & 0.6825 & 0.6714 & 0.6769 & 20 \\
& NLL & 0.8879 & 0.5730 & 0.6968 & 36\\

\midrule[1pt]

\end{tabular}
\flushleft{CE: cross entropy loss; Focal: Focal loss; NLL: negative log-likelyhood loss}

\end{table*}

\section{Experiments and Evaluation}\label{section:sec4}
\subsection{Datasets}
We use the supply chain credit data from \cite{shi2023transcoralnet}, which consists of four training datasets (owned by four clients) and an independent testing dataset. The number of samples in the four training datasets is 18,368, 19,904, 18,816, and 13,440, respectively. Each client dataset comprises different borrower datasets from various supply chain industries. The defaulting (minority class) rates for the clients are 11.75\%, 12.45\%, 14.04\%, and 13.52\%, respectively. The testing dataset consists of data from the four clients, each contributing 20\% of their respective datasets. The size of the testing dataset is 17,664. The four client datasets are described by the same 21 features \cite{shi2023transcoralnet}.

\subsection{Baseline Comparisons:}
We compared our proposed method with two benchmark models: FedProx \cite{li2020federated} and FedAvg \cite{mcmahan2017communication}. These models adopt a weighted aggregation and random client selection strategy. Each benchmark model has the same local model architecture as the proposed model, which consists of a transformer block and fully connected layers. Each model is tested under three different loss functions: Cross Entropy Loss, Focal Loss, and Negative Log-Likelihood Loss. These three loss functions share the same weights.

\subsection{Experimental Setup}
To choose hyperparameters affecting the model's performance, we adopt a stratified dataset split to preserve the proportion of defaulting and non-defaulting samples within each client dataset. Specifically, for each client, 80\% of the local dataset is used for training and 20\% for validation. This ensures balanced representation of minority and majority classes across both splits, mitigating potential bias from uneven distributions.. Empirical study suggest that models often achieve sufficient accuracy after 50 rounds of training \cite{zhu2021federated}. We set the number of communication rounds to 50 and select the best-performing model as the final one. The regularization parameter of the proximal term, denoted as $\mu$, increases from 0 to 0.01 over 50 communication rounds in steps of 0.0002. This communication-round-varying $\mu$, which changes with the number of rounds, helps the local model reach optimal performance initially and allows the global model to converge more quickly in the later stages of training. The selection ratio of clients participating in training, $r$, is set to 0.5. We set the $\gamma_k$ value based on the proportion of data volume from each client selected in each communication round, and the sum of all $\gamma _k $ equals 1. Local models are trained for 250 epochs using the Stochastic Gradient Descent (SGD) optimizer with a batch size 64 and a fixed learning rate of 0.01. The weights of the loss function are empirically determined to be 0.25 and 0.75 for the majority and minority classes, respectively. Additionally, we use three attention heads in the transformer encoder block. Our model implementation is based on Python 3.10, PyTorch 1.9.0, and Tenseal 0.3.14.

\subsection{Model evaluation}
We chose the final model, which is the global model with the best performance in 50 communication rounds. Borrowers who default can directly cause financial institutions to experience profit losses. The repercussions of incorrectly identifying defaulters are considerably more severe than those of incorrectly identifying non-defaulters in credit assessment model predictions. Therefore, we rely on positive class recall as our primary model evaluation indicator. Conversely, incorrectly classifying a regular borrower as a defaulter can result in customer attrition. In such instances, we employ the F1 score as a balanced solution. Both recall, and F1 scores are computed based on positive data.

\begin{figure*}[htp]
\centering
\subfloat[]
{\includegraphics[width=14.5cm]{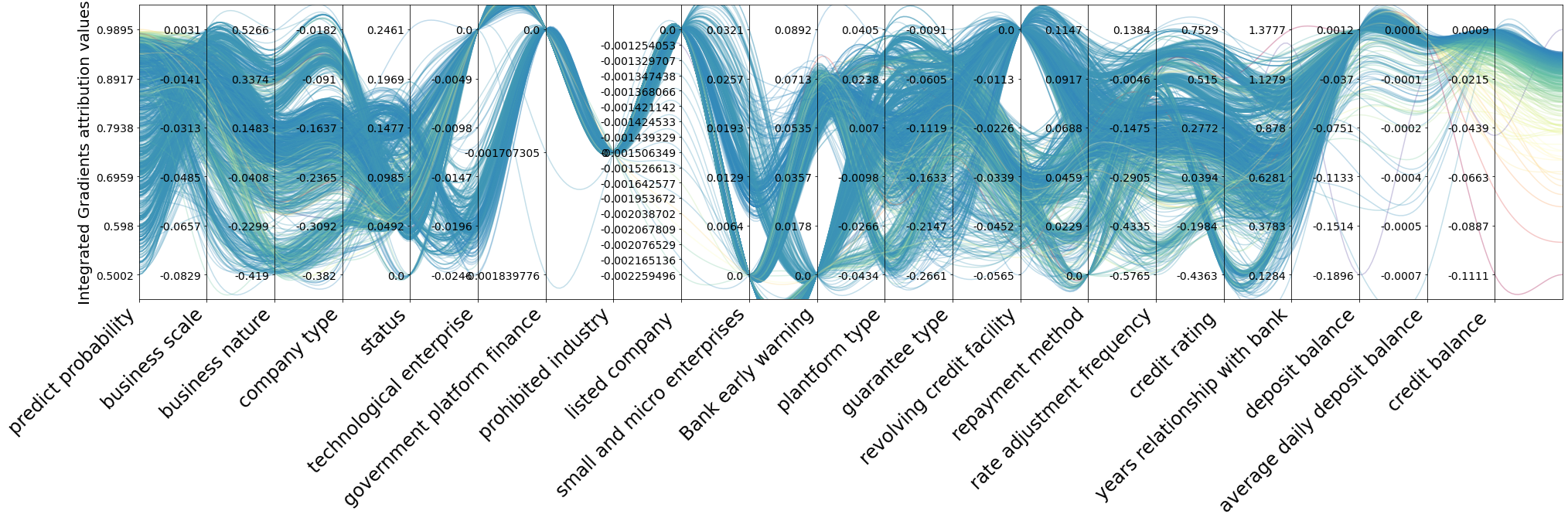}\label{fig2:f1}}
\hfill
\subfloat[]
{\includegraphics[width=14.5cm]{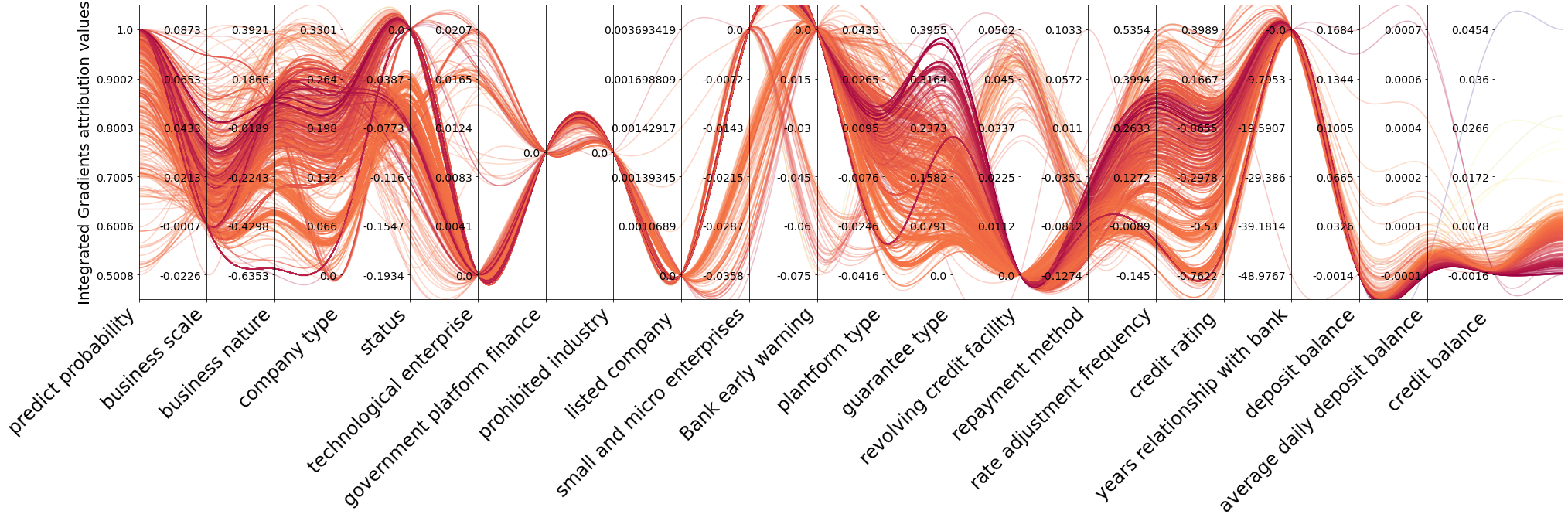}\label{fig2:f2}}

\caption{Integrated Gradients–based explanation results for defaulting and non-defaulting client samples. 2,000 representative samples from Client 1 were randomly selected for each group. The vertical axes represent the attribution values derived from the Integrated Gradients method, indicating the relative contribution of each input feature to the model’s predicted probability of default. The horizontal axes list the company attributes and financial indicators.  (a) Integrated gradients results for defaulting samples. (b) Integrated gradients results for non-defaulting samples.enterprises.}\label{fig2}
\end{figure*}

\begin{figure*}[htpb]
 \centering
 \subfloat[]{\includegraphics[width=7.5cm]{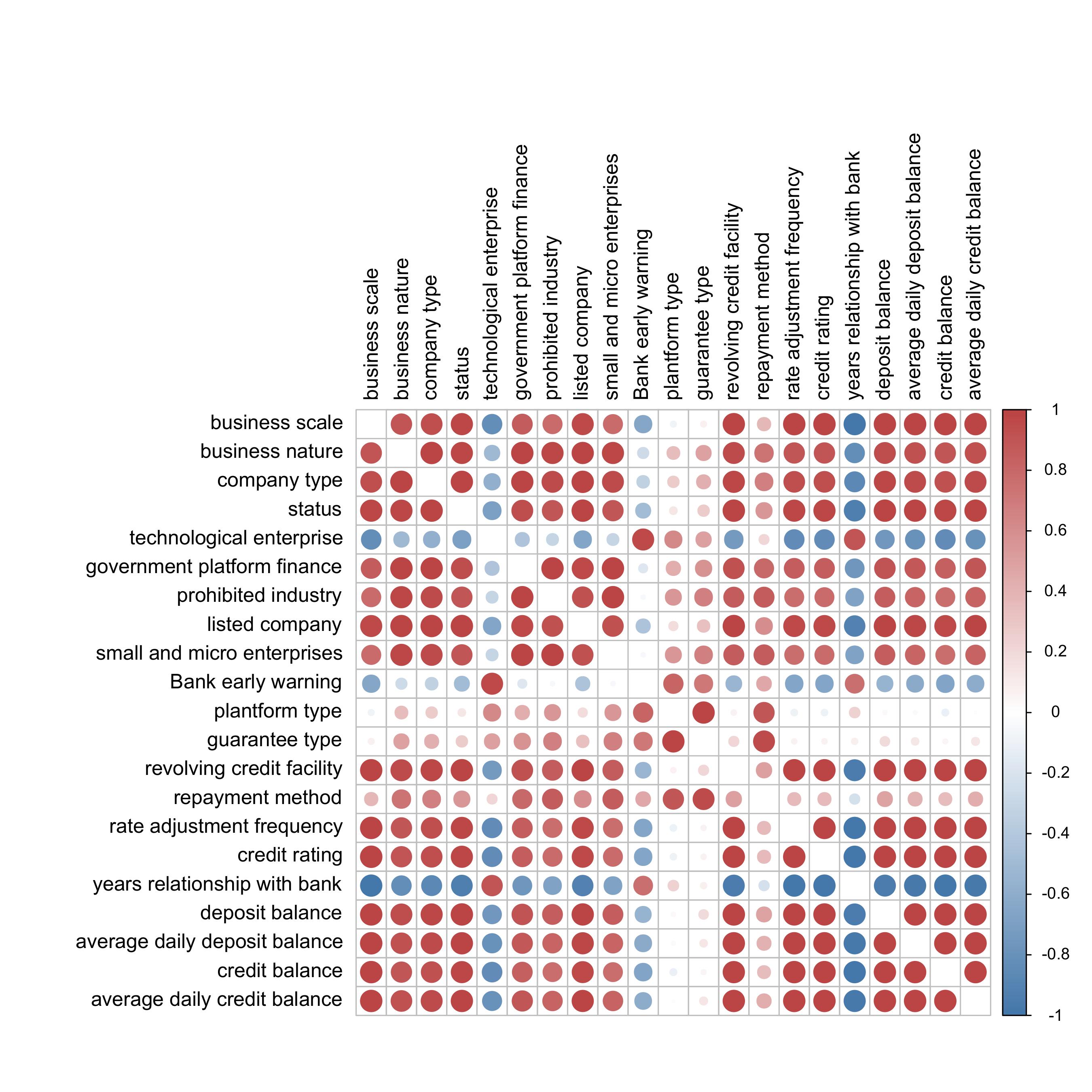}\label{fig3:f1}}
 \hfill
 \subfloat[]{\includegraphics[width=7.5cm]{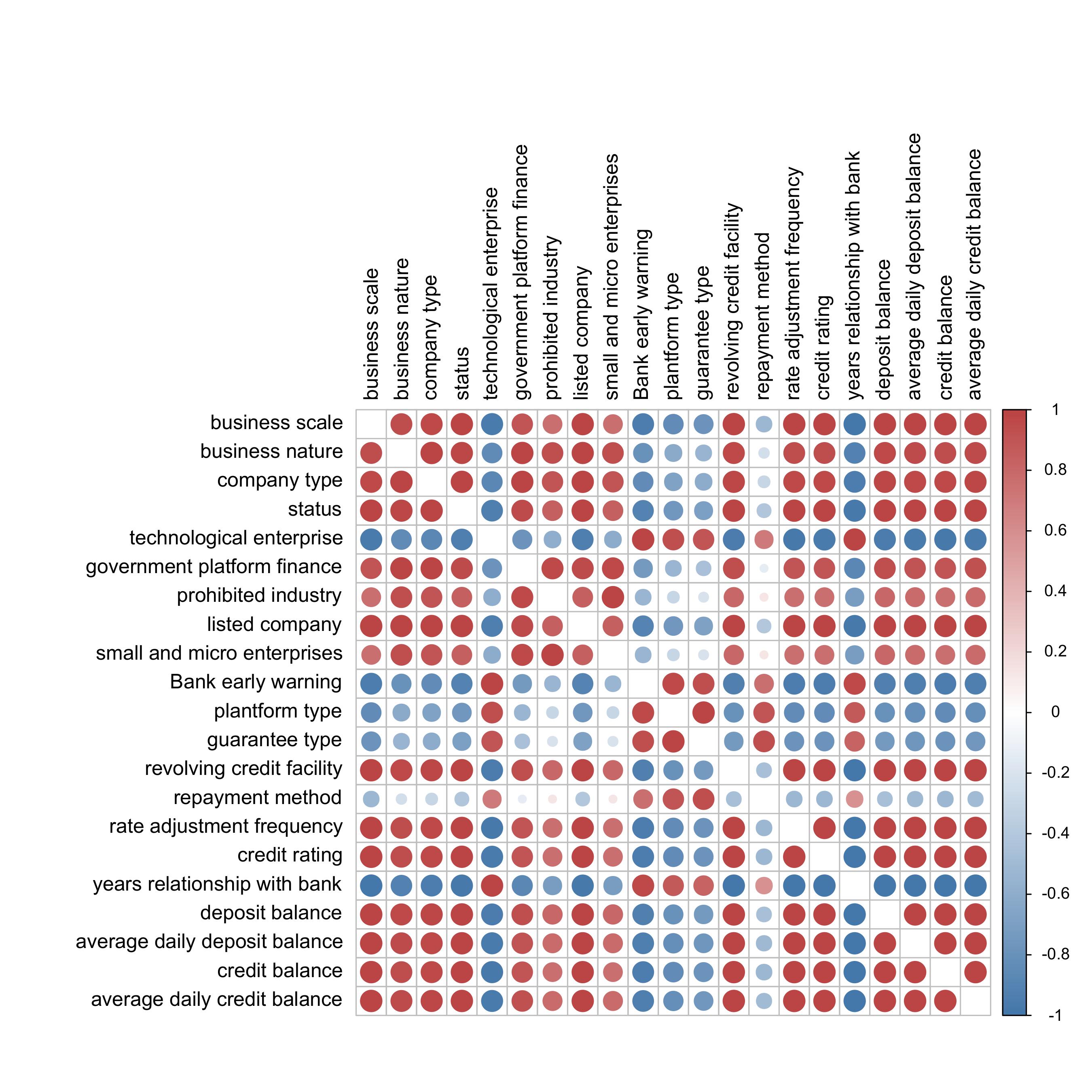}\label{fig3:f2}}

 \caption{Visualization of attention scores for defaulting and non-defaulting samples. To provide an interpretable view of how the model allocates attention across input features, the figure shows pairwise correlations of feature-level attention weights in a matrix form. The size of each circle reflects the magnitude of the attention score, while the color represents the sign and intensity of the relationship (red for positive, blue for negative).  (a) Attention scores for defaulting samples. (b) Attention scores for non-defaulting samples. }\label{fig3}
\end{figure*}

\section{Results and discussion}\label{section:sec5}
\subsection{Overall performance}
The performance of our proposed Trans-XFed
model and other tested approaches are depicted in Table \ref{tbl1}. The table provides the best recall, precision, and F1 scores of the minority class (defaulting) on testing datasets from 50 communication rounds using different federated learning models and loss functions. The communication round column indicates the round in which the best model appears. 
Trans-XFed with the NLL loss achieved the highest recall (0.8889) and F1 score (0.7137) in just 10 communication rounds, outperforming other methods in speed and balance between precision and recall. While FedAvg with NLL achieved a similar recall (0.8879), it required 36 rounds. These results highlight Trans-XFed's efficiency and effectiveness.

Trans-XFed outperforms FedProx under the same framework for three main reasons. The first reason is the communication-round-varying $\mu$, which allows the local model to focus on its training in the early stages, improving the effectiveness of the local model. As the parameters $\mu$ increase, the local model does not deviate from the global model in the later stages. The second reason is the client selection strategy, which selects clients with highest F1 scores in each round to participate in training and parameter aggregation, thereby ensuring the effectiveness of the global model. The third reason is the use of a transformer-based architecture, which could capturing complex, hierarchical relationships in data, leading to better feature representation and explanation.

\subsection{Interpretation}
The Integrated Gradients analysis identifies which features most influence whether a borrower is predicted to default. Each feature receives an attribution score: positive scores indicate a positive contribution to default prediction, negative scores the opposite, and the magnitude reflects contribution strength.

Figure~\ref{fig2} compares feature contributions for defaulting and non-defaulting samples. For defaults, features such as \textit{‘small and micro enterprises’}, \textit{‘bank early warning’}, \textit{‘revolving credit facility’}, and \textit{‘years relationship with bank’} contribute strongly and positively, indicating higher risk. For non-defaults, \textit{‘guarantee type’} and \textit{‘credit rating’} are most influential, typically reducing default likelihood when stronger. Factors such as \textit{‘government platform finance’} and \textit{‘prohibited industry’} show minimal effect in both groups.

This analysis gives decision makers useful insights. They can focus on borrowers with weak credit guarantees, years relationships with the bank, or heavy use of revolving credit when assessing risks. Taking actions like adjusting loan terms can help lower the chances of default for these borrowers.

\subsection{Attention analysis}
An attention score quantifies the importance of one feature relative to another feature or a set of features. The higher attention value indicates a strong relationship between the two features. We calculate the average attention scores of defaulting and non-defaulting samples, respectively. The heat-maps (Figure \ref{fig3}) display the attention scores normalized in the range $[-1,1]$ between pairs of features for defaulting samples (Figure \ref{fig3:f1}) and non-defaulting samples (Figure \ref{fig3:f2}). The heat-maps effectively visualize the relationships between features for defaulting and non-defaulting samples. The differences in these relationships highlight the distinct feature interactions that could be crucial for understanding and predicting default behavior. By analyzing these attention scores, one can gain insights into which features are most influential in distinguishing between defaulting and non-defaulting samples, aiding in developing better predictive models.

For defaults, \textit{‘repayment method’} shows consistently strong positive relationships with most features, making it central to default prediction. In non-defaults, its influence is mixed, depending on borrower profile. \textit{‘Platform type’} and \textit{‘guarantee type’} show weaker links in defaults but often negative relationships in non-defaults, suggesting a mitigating effect against default risk.

\subsection{Computational cost and scalability} One key advantage of the proposed Trans-XFed framework lies in its computational efficiency and scalability across distributed clients. Compared to baseline federated learning methods such as FedAvg and FedProx, Trans-XFed achieves faster convergence with fewer communication rounds, significantly reducing the overall training time. For instance, as shown in Table I, Trans-XFed reaches its best performance at round 10, whereas FedAvg and FedProx typically require more than 30 rounds. This efficiency is primarily driven by our performance-based client selection strategy, which prioritizes clients with higher-quality updates, allowing the global model to learn more effectively from each round.

Local training costs are kept manageable through a lightweight transformer encoder with three attention heads and shallow depth. Training is fully parallelized across clients, enabling the framework to scale to larger networks without significant bottlenecks. While CKKS homomorphic encryption introduces some overhead during parameter exchange, the faster convergence offsets this cost, resulting in efficient and privacy-preserving communication.  
Overall, the combination of selective client participation, parallelizable local computation, and early convergence makes Trans-XFed a practical and scalable solution for real-world federated credit assessment tasks.

\section{Conclusion}\label{section:sec6}
In this paper, we introduced Trans-XFed, an explainable federated learning framework designed for supply chain credit assessment. Our approach combines the robustness of the FedProx architecture with the powerful clients selection strategy, feature extraction capabilities of transformer models, and integrated gradients explanation approach, ensuring effective handling of heterogeneity and explainablity. A particularly notable advantage of our method is the significant reduction in computational cost. The performance-based client selection strategy allows the algorithm to converge much faster compared to baseline approaches. The explainability aspect of our model provides valuable insights into the decision-making process, addressing one of the key challenges in the financial industry.  Moreover, the fully homomophic encryption ensure the model parameters exchange under privacy and secure.  Our results indicate that Trans-XFed not only improves predictive performance but also offers a practical solution for real-world applications where data privacy and security are paramount.

\bibliographystyle{ieeetr}

\end{document}